\documentclass[11pt,draftcls,onecolumn]{IEEEtran}%
\usepackage{yhmath}
\usepackage{graphics}
\usepackage{textcomp}
\usepackage{amssymb}
\usepackage{amscd}
\usepackage{dsfont}
\usepackage{amsmath}
\usepackage{epsfig}
\usepackage{psfrag}
\usepackage{rotating}
\usepackage{amsmath}
\usepackage{amsfonts}
\usepackage{url}
\usepackage{color}

\newtheorem{theorem}{Theorem}

\newtheorem{definition}[theorem]{Definition}

\newcommand{\expected}[2]{\mathds{E}_{#2}\left[ #1 \right]}

\newcommand{\bs}{\boldsymbol}

\newcommand{\argmax}{\mathrm{argmax}}

\newcounter{MythCounter}
\begin{document}
\title{Learning Equilibria with Partial Information in Decentralized Wireless Networks}
\author{L.~Rose, 
        S.~M.~Perlaza, 
        S. Lasaulce 
        ~and~M.~Debbah 
\thanks{L. Rose is with Thales Communication, 160 Boulevard de Valmy, 92700 Colombes, France (e-mail: luca.rose@fr.thalesgroup.com)}
\thanks{S. M. Perlaza is with the Alcatel-Lucent Chair in Flexible Radio at SUPELEC. $3$ rue Joliot-Curie, $91192$, Gif-sur-Yvette, cedex. France. (samir.medinaperlaza@supelec.fr)}%
\thanks{S. Lasaulce is with the Laboratoire des Signaux et Syst\`{e}mes (LSS) at SUPELEC. $3$ rue Joliot-Curie, $91192$, Gif-sur-Yvette, cedex. France. (samson.lasaulce@lss.supelec.fr)}
\thanks{M. Debbah is with the Alcatel-Lucent Chair in Flexible Radio at SUPELEC. $3$ rue Joliot-Curie, $91192$, Gif-sur-Yvette, cedex. France. (merouane.debbah@supelec.fr)}%
}

\maketitle

\begin{abstract}
In this article, a survey of several important equilibrium concepts for decentralized networks is presented. The term decentralized is used here to refer to scenarios where decisions (e.g., choosing a power allocation policy) are taken autonomously by devices interacting with each other (e.g., through mutual interference). The iterative long-term interaction is characterized by stable points of the wireless network called equilibria. The interest in these equilibria stems from the relevance of network stability and the fact that they can be achieved by letting radio devices to repeatedly interact over time. To achieve these equilibria, several learning techniques, namely, the best response dynamics, fictitious play, smoothed fictitious play, reinforcement learning algorithms, and regret matching, are discussed in terms of information requirements and convergence properties. Most of the notions introduced here, for both equilibria and learning schemes, are illustrated by a simple case study, namely, an interference channel with two transmitter-receiver pairs.

\end{abstract}

\section{Introduction}\label{SecIntroduction}

The notion of cognitive radio (CR) has gained momentum in recent years to build flexible and efficient networks. Indeed, CRs are nowadays widely accepted as a suitable solution to rationally exploit shared spectral resources and increase spectral efficiency. 
The main idea behind CR relies on the capability of a given radio device to self-configure its own communication parameters 
in an intelligent, autonomous and decentralized manner, as a result of its interaction with the environment. In this context, the choice of a particular communication configuration by a given CR is highly influenced by the choice of all other radio devices.
Within this framework, non-cooperative game theory appears as a suitable paradigm to study and analyse such scenarios. Therefore, the idea of equilibrium, namely, Nash equilibrium (NE), becomes particularly relevant. Indeed, at the NE, the transmit configuration of each CR in the network is optimal with respect to the configuration of all its counterparts. Interestingly, in some cases, an equilibrium can be reached by using particular iterative procedures similar to learning processes \cite{Fudenberg-98}. 

\noindent
In this article, we first present an overview of various equilibrium concepts. Later, we introduce a set of learning algorithms particularly relevant to achieving equilibrium in wireless networks. For each algorithm, we discuss the required information that each CR must possess at each iteration and the convergence properties.
 
\noindent
The rest of the paper is organized as follows.
In Sec. \ref{def_abb}, we briefly present the notations adopted in this paper, as well as usual game-theoretic terminology. In Sec. \ref{eq_conc}, we present and discuss several important solution concepts for games, namely, the coarse correlated equilibrium, the correlated equilibrium, the Nash equilibrium, and the $\epsilon-$Nash equilibrium. In Sec. \ref{learning}, we discuss important learning algorithms which can, under certain conditions, converge to one of the aforementioned solutions. In Sec. \ref{IC}, we present an illustrative case study: the $2 \times 2$ interference channel, a simple, though very important, communication scenario, which we use as a test-bench for comparing the aforementioned algorithms.

\section{Definitions and abbreviations}\label{def_abb}

In game theory, the normal form is a convenient mathematical representation of a game. Basically, it consists of a triplet: the set of players $\mathcal{K}=\lbrace 1,2,...,K \rbrace$, the set of actions $\mathcal{A}_k=\lbrace A_k^{(1)},...,A_k^{(N_{k})} \rbrace$, $\forall k\in\mathcal{K}$, and the utility functions $u_k(\bs{a})$, where $\bs{a} \in \mathcal{A} = \mathcal{A}_1\times\mathcal{A}_2\times...\times\mathcal{A}_K$ is an action profile/vector. With a
 slight abuse of notation, we denote by $\bs{a}_{-k} \in \mathcal{A}_{-k}$ the vector of actions of all players except the $k$-th player and we write the vector $\bs{a}$ as $(a_k,\bs{a}_{-k})$ to stress the $k$-th component. For instance, the set of players can consist of the set of wireless terminals present in the network, the action set can be any feasible vector of transmit powers, and the utility function can be the spectral efficiency. Other components are also possible for the game representation and they depend on the scope and purpose of the network design.
We denote by $\triangle\left( \mathcal{A}\right)$ the set of all possible probability distributions over the whole set of actions $\mathcal{A}$, and by $\triangle(\mathcal{A}_k)$ the set of all possible probability distributions of user $k$ over its action set. We refer to the elements of the set $\mathcal{A}_k$ as \textit{actions} of player $k$ and those of the set $\triangle(\mathcal{A}_k)$ as \textit{strategies} of player $k$. A given strategy of player $k$ is denoted by $\bs{\pi_{k}}=(\pi_{k,A_k^{(1)}},...,\pi_{k,A_k^{(N_{k})}}) \in \triangle\left(\mathcal{A}_k\right)$, where $\pi_{k,A_k^{(n_{k})}}$ represents the probability that player $k$ plays action $A_K^{(n_{k})}$. We indicate by $\bs{\phi}=(\phi_{A^{(1)}},...,\phi_{A^{(N)}}) \in \triangle\left( \mathcal{A} \right)$, with $N=\prod_{j=1}^{K}N_j$, a given joint probability distribution over the set $\mathcal{A}$, with $\phi_{A^{(n)}}$ being the probability of observing $A^{(n)}$ as an outcome of the game.  

\section{From coarse correlated equilibria to Nash equilibria}
\label{eq_conc}

The most general type of equilibria we use in this paper is the
so called \emph{coarse correlated equilibrium} (CCE) \cite{Young-2004}. The
idea behind CCE is that actions chosen by the players of a game may be correlated. For instance, correlation may appear when a common
broadcast signal is observed by several transmitters choosing their
transmit configuration, e.g., a power control policy. We call the signals
received by the players recommendations. In such a context, a CCE is a probability distribution $\bs{\phi} \in \triangle\left( \mathcal{A} \right)$
over the set of action profiles of the game from which no player has
interest 
in unilaterally deviating. The realizations of this joint distribution $\bs{\phi} $ are the recommendations. Mathematically, this can be written as follows.
\begin{definition}[Coarse Correlated Equilibrium]\label{DefCCE}\emph{A
joint probability distribution $\bs{\phi} \in \triangle\left(\mathcal{A}\right)$ is a CCE if
$\forall k \in\mathcal{K}$ and $\forall a'_{k} \in \mathcal{A}_k$ it holds that
\begin{equation}\label{CCE_eq}
\sum_{\bs{a}\in\mathcal{A}}u_k(\bs{a})\phi_{\bs{a}}
\geq \sum_{a_{-k}\in\mathcal{A}_{-k}}u_{k}(a'_{k},\bs{a}_{-k})\phi_{-k,\bs{a}_{-k}},
\end{equation}
where $\phi_{-k,a_{-k}}=\sum_{a_k \in \mathcal{A}_k} \bs{\phi}_{\left(a_k,\bs{a}_{-k}\right)}$ is the marginal probability distribution w.r.t. $a_k$. 
}
\end{definition}

\noindent
An important remark is that, following the notion of CCE, players are assumed to decide, \emph{before} receiving the recommendation, whether to commit to follow it or not. At a CCE, all players are willing to commit to follow the recommendation given that all the others also choose to commit. That is, if a single player decides not to commit to follow the recommendations, it experiences a lower (expected) utility. 
A special case of CCE is the \emph{correlated equilibrium} (CE, \cite{Young-2004}). The difference between the CCE and the CE is that, in the latter, players choose whether to follow or not a given recommendation, \emph{after} it has been received.  Therefore, there is not \textit{a priori} commitment. It follows, in particular, that every CE is a CCE \cite{Young-2004}.

\noindent
Now, if the players choose their strategy following independent individual probability distributions $\bs{\pi}_k \in \triangle\left( \mathcal{A}_k \right)$, i.e., $\phi_{a} = \prod_{j=1}^{K} \pi_{j,a_j}$ in \eqref{CCE_eq}, we obtain from Def. \ref{DefCCE}, the definition of \emph{mixed Nash equilibrium} (MNE); the MNE is, clearly, a special case of CE and thus, a special case of CCE. A MNE is, therefore, a vector of
individual probability distributions $\bs{\pi} = \left(\bs{\pi}_1, \ldots, \bs{\pi}_K \right)$ which is stable to unilateral deviations, i.e., if player $k$ decides to use a different probability distribution from the corresponding $\bs{\pi}_k$, then it observes a lower (expected) utility.
As shown in \cite{Fudenberg-Tirole-1991}, this type of equilibria always exists in games with
finite number of players and finite action sets. For more results on the existence and multiplicity of NE, the reader is referred to \cite{Lasaulce-Tutorial-09}. The finiteness
assumption is especially relevant when a wireless terminal has to select a given communication setting, e.g., a channel, a constellation size, or a transmit power level.

\noindent
We further introduce the concept of \emph{pure NE} (PNE). A PNE is obtained by restricting the players to deterministically choose one of their actions instead of choosing it by following a probability distribution. A PNE is therefore a special
case of MNE where the individual probability distribution is a Dirac delta function over
a given action. Thus, a PNE is a vector of actions $\bs{a} = \left( a_1, \ldots, a_K \right)$
stable to unilateral deviations, i.e., if player $k$ uses a different action from its corresponding $a_k$, while the others keep their equilibrium action, player $k$ observes a lower (instantaneous) utility.

\noindent
As a last notion of equilibrium, we introduce the idea of
$\epsilon-$\emph{equilibrium} \cite{Young-2004}. An $\epsilon-$equilibrium is
a mixed strategy profile $\bs{\pi} = \left(\bs{\pi}_1, \ldots, \bs{\pi}_K \right)\in\triangle\left( \mathcal{A}_1\right)\times\ldots\times\triangle\left( \mathcal{A}_K\right)$ such that if only one player $k$ uses a different strategy from its corresponding $\bs{\pi}_k$, it does not observe a utility improvement greater than $\epsilon>0$. An instance of $\epsilon-$NE \cite{Young-2004} is the logit equilibrium \cite{Young-2004}. In what follows, some learning algorithms which may converge to CCE, CE, MNE, PNE, or $\epsilon-$equilibrium are provided.

\section{Learning Equilibria}\label{learning}
The process of learning equilibria is basically an iterative process. Each iteration of the learning process can be broadly divided into three  phases: $(i)$ the observation of the environment at iteration $n$, which gives an idea to the players how well they played in the previous iteration; $(ii)$ the improvement of the strategy $\bs{\pi}_k(n)$ based on the current observation and $(iii)$ the selection of the action $a_k(n)$ according to the strategy $\bs{\pi}_k(n)$. Hence, we say that players learn to play an equilibrium, if after a given number of iterations, the strategy profile $\bs{\pi}(n) = \left(\bs{\pi}_1(n), \ldots, \bs{\pi}_{K} (n)\right) \in \triangle\left( \mathcal{A}_1\right)\times\ldots\times \triangle\left( \mathcal{A}_K\right)$ converges to an equilibrium strategy.

\noindent
The purpose of this section and Sec. \ref{IC} is, under the space limitations for this survey, to introduce the following set of learning algorithms: best response dynamics (BRD), fictitious play (FP), smooth fictitious play (SFP), regret matching (RM), reinforcement learning (RL) and the \emph{joint utility and strategy learning estimation} reinforcement learning (JUSTE-RL).
Then, in Sec. \ref{sec:discussion}, we compare such algorithms in terms of relevant features in the context of wireless communications, for instance, type of observations, type of action sets, convergence time, nature of the steady state achieved when convergence is observed and conditions for convergence.

\subsection{Informal definition of the learning algorithms under consideration}
\label{sec:algo-definitions}

\subsubsection{Best Response Dynamics} 
In its most basic form, the best response dynamics relies on the following assumption: at each game stage $n \in \mathds{N}$, every player $k$ plays the action $a_k(n)$ which optimizes its own utility given the actions played by the other players. When all players play simultaneously at each stage (simultaneous-BRD), the optimization of player $k$ is done with respect to the action profile $\bs{a}_{-k}(n-1)$. When players play sequentially, only one player at each stage (sequential-BRD) updates its action $a_k(n)$, optimizing it with respect to the action profile $\left(a_{1}(n),\ldots, a_{k-1}(n), a_{k+1}(n-1),\ldots, a_{K}(n-1)\right)$.
%

\subsubsection{Fictitious Play} 
The fictitious play relies on the assumption that at each stage $n$, each player $k$ knows all the past actions of all the other players, i.e., $a_j(0), \ldots, a_j(n-1)$, $\forall j \in \mathcal{K}\setminus\lbrace k \rbrace$. Based on such observations, player $k$ calculates the empirical frequencies with which each player plays its corresponding actions. We refer to these empirical frequencies as beliefs.  Let us denote the belief that player $k \neq j$ has on player $j$ by the vector $\bs{f}_j(n) = \left(f_{j,A_j^{(1)}}(n), \ldots, f_{j,A_j^{(N_j)}}(n)\right) \in \triangle\left( \mathcal{A}_j\right)$. 
At each stage, all players (simultaneously or sequentially, as in the BRD)
choose their current action by optimizing their expected utility with respect to the beliefs on all the other players, i.e., $a_k(n) \in \argmax_{a_k \in \mathcal{A}_k} \expected{u_k\left(a_k,\bs{a}_{-k}\right)}{\bs{f}(n)}$, where $\bs{f}(n) = \left(\bs{f}_{1}(n), \ldots, \bs{f}_{K}(n)\right)$.

\subsubsection{Smooth Fictitious Play (SFP)} 
The convergence of FP is not ensured in games with cycles and its ability to explore the whole action set is highly constrained. To overcome these issues, a simple variation of the FP has been proposed under the name of smooth fictitious play (SFP). The assumptions on which SFP relies are the same as FP and actions can be updated either simultaneously or sequentially. The main difference between SFP and FP is that, at each stage $n$, player $k$ does not choose a deterministic action. It rather builds a probability distribution $\bs{\pi}_k(n) \in \triangle\left( \mathcal{A}_k \right)$ to choose its action $a_k(n)$. Such a probability distribution can be interpreted as the one that maximizes a weighted sum of the original expected utility and other continuous strictly concave function. For instance, if such a function is the \emph{entropy function} \cite{Young-2004}, the resulting probability distribution is given by the logit probability distribution.
%
%

\subsubsection{Regret Matching (RM)} 
Contrary to the case of BRD, FP and SFP, where players determine whether to play or not a particular action based on the idea of utility maximization, in RM, such a decision is made considering the notion of regret minimization.
The regret that player $k$ associates with action $A_k^{(n_k)}$ is defined as the difference between the average utility the player would have obtained by always playing $A_k^{(n_k)}$ and the average utility actually achieved with the current strategy, i.e.,
\begin{equation}
r_{k,A_k^{(n_k)}}(n)=\frac{1}{n-1}\sum_{t = 1}^{n-1}(u_k(A_k^{(n_k)},\bs{a}_{-k}(t)) - u_k(a_k(t), \bs{a}_{-k}(t))).
\label{regr}
\end{equation}
RM relies on the assumptions that at every stage $n$, player $k$ is able to both evaluate its own utility, i.e., to calculate $u_k(a_k(n),\bs{a}_{-k}(n))$ and compute the utility it would have obtained if it had played any other action $a_k'$, i.e. $u_k(a_k',\bs{a}_{-k}(n))$. 
Finally, the action to be played at stage $n$ is taken following the probability distribution $\bs{\pi}_k(n)$, which is obtained by normalizing to one the regret vector $\bs{r}_k(n) = \left( r_{k,A_k^{(1)}}(n), \ldots, r_{k,A_k^{(N_k)}}(n)\right)$.
%


\subsubsection{Reinforcement Learning (RL)} 
In the case of reinforcement learning (RL), players are modelled as automata that implement a given behavioural rule. 
In general, RL techniques rely on the following two conditions: $(i)$ for each player $k$, the action set $\mathcal{A}_k$ is finite and for all action profiles $\bs{a} \in \mathcal{A}$, the achieved utility $u_k\left( a_k, \bs{a}_{-k}\right)$ is bounded; $(ii)$ each player is able to periodically observe its own achieved utility.  Intuitively, the idea behind CRL is that  actions leading to higher utility observations in stage $n$ are granted with higher probabilities in the game stage $n+1$, and vice versa.
%
%

\subsubsection{Joint Utility and Strategy Estimation based - Reinforcement Learning (JUSTE-RL)} A variant of the former algorithm has been proposed in \cite{Perlaza-Spawc2010}. The joint utility and strategy estimation behavioural rule relies on the same assumptions as the classical RL. The main difference between classical RL and JUSTE-RL is that, in the former, the observation $\tilde{u}_k(n)$ of the utility of player $k$ is used to directly modify the probability distribution $\bs{\pi}_k(n)$; in the latter, such an observation is used to build an estimation of the expected utility for each of the actions. Such utility estimates are, then, used in the same iteration to finally build a probability distribution $\bs{\pi}_k(n)$ from which action $a_k(n)$ will be drawn. Thus, each player always possesses an estimation of the expected utility it obtains by playing each of its actions.  

\subsection{Discussion}\label{sec:discussion}

The purpose of this section is to provide additional insights about the performance and pertinence of the learning algorithms described above in the context of decentralized wireless networks. In the following, we compare the algorithms in terms of several fundamental features. We summarize this discussion in Table \ref{FigTable}.

\subsubsection{Observations}

At each iteration of a given learning algorithm, each player must obtain some information about how the other players are reacting to its current action, in order to update their strategy and choose the following action. Broadly speaking, in algorithms such as BRD, FP, SFP and RM, players must observe the actions played by all the other players. This implies that a large amount of additional signaling is required to broadcast such information in wireless networks. In some particular cases, this condition can be relaxed and less information is required \cite{Larsson-2009, Leshem-Zehavi-2009}. However, this is highly dependent on the topology of the network and the explicit form of the utility function \cite{Scutari-Algorithms-2008}.
Other algorithms, such as RL and JUSTE-RL, only require that each player observes its corresponding achieved utility at each iteration. This is in fact, their main advantage, since such information requires a simple feedback message from the receiver to the corresponding transmitters  \cite{sastry-1994, Perlaza-Spawc2010}.
%

\subsubsection{Knowledge and Calculation Capabilities}
Learning algorithms such as BRD, FP, SFP and RM involve an optimization problem at each iteration \cite{Fudenberg-98}, that is, either the maximization of the (expected or instantaneous) utility or minimization of the regret. Therefore, generally, highly demanding calculation capabilities are required to implement them. More importantly, solving such optimization requires the knowledge of a closed-form expression of the utility function. This implies that each player knows the structure of the game, i.e., set of players, action sets, current strategies, channel realizations, etc.  
In this respect, RL and JUSTE-RL algorithms are more attractive since  only algebraic operations are required to update the strategies. In terms of knowledge, in both RL and JUSTE-RL, players are only required to know,  at each iteration, the action they actually played and the corresponding achieved utility. Indeed, one can say that players are not even aware of the presence of other players. 

\subsubsection{Nature of the Action Sets}

The nature of the action sets of the game plays an important role. The BRD can be used for both continuous and discrete action sets, whereas in their standard versions FP, SFP, RM, CRL, and JUSTE-RL are designed for discrete action sets. For instance, action sets are discrete in problems where a channel, constellation size or discrete power levels must be selected, whereas continuous sets are more common in power allocation problems \cite{Lasaulce-Tutorial-09}. 

\subsubsection{Steady State}

When a steady state is achieved by one of the algorithms under consideration, such state may correspond to one of the equilibrium notions presented in Sec. \ref{eq_conc}. In particular, when BRD and FP converge, the strategy of the players at the steady state is a NE. In the case of the RM, it converges to an element of the set of CCE. Here, we highlight the fact that, even though the notion of CCE relies on the idea of the recommendations studied in Sec. \ref{SecIntroduction}, it does not require the existence of recommendations to converge to an element of the set of CCE. 
When SFP or JUSTE-RL achieve a steady state, it corresponds to an $\epsilon$-NE. On the contrary, in the case of RL, a steady state not necessarily corresponds to a particular notion of equilibrium.

\subsubsection{Convergence Conditions}

Regarding the conditions for convergence, only sufficient conditions are available. As shown in Table \ref{FigTable}, the considered algorithms typically converge in certain classes of games \cite{Young-2004} such as dominance solvable games (DSG), potential games (PG), super-modular games (SMG),  $2\times N$ non-degenerated games (NDG) or zero-sum games (ZSG). 

\subsubsection{Synchronization}

In the particular case of algorithms where each player must observe the actions of the others, e.g., BRD, FP, SFP and RM, certain synchronization is required in order to allow players to know when to play and when to observe the actions of the others. In wireless communications, this requirement implies the existence of a given protocol for signalling messages exchange. 
 Conversely, when players require only an observation of their individual utility, such a synchronization between all the players becomes irrelevant. Here, only a feedback message from the receiver to the corresponding transmitters per learning iteration is sufficient.
 
\subsubsection{Environment}

Learning techniques such as the BRD are highly constrained for real system implementations since they require the network to be static during the whole learning processes. On the contrary, all the other techniques allow the dynamics of the network to be captured by their statistics as long as they are stationary. This is basically because, contrary to BRD, all the other techniques determine whether to play or not a particular action based on the expected utility rather than instantaneous utility. 

\subsubsection{Convergence Speed}

The speed of convergence (when it is observed) is highly influenced by the amount of information available for the players. For instance, FP, SFP and RM converge faster than JUSTE-RL since the formers calculate the expected utility relaying on a closed form expression. Conversely, the latter calculates it as the time-average of the instantaneous observations of the achieved utility. This requires a large number of observations to obtain a reliable approximation to the expected utility. We do not state any particular comment on the speed of convergence of BRD and RL since, in the former the scenario is considered fixed and the latter, it does not necessarily converge to an equilibrium strategy. However, conclusions for a particular case are stated in the following section.

\section{Case Study: The Parallel Interference Channel}\label{IC}

In this section, we introduce a simple but insightful  example, which we use as a test-bench to compare the learning algorithms described above. Consider a parallel interference channel, that is, a set of $2$ transmitter-receiver pairs sharing a set of $S$ non-overlapping frequency bands. For the ease of presentation, assume that channel gains are time invariant during the whole transmission duration. Each transmitter chooses a single frequency band to transmit aiming to maximize its individual spectral efficiency, i.e., the ratio between the individual Shannon rate and available bandwidth. This problem has been analysed in the context of compact and convex sets of actions in \cite{Scutari-Algorithms-2008} and in discrete and finite sets in \cite{Rose-Perlaza-2011}, which is the case of this section.

\begin{figure}[t]
\centering
\includegraphics[width=1\textwidth]{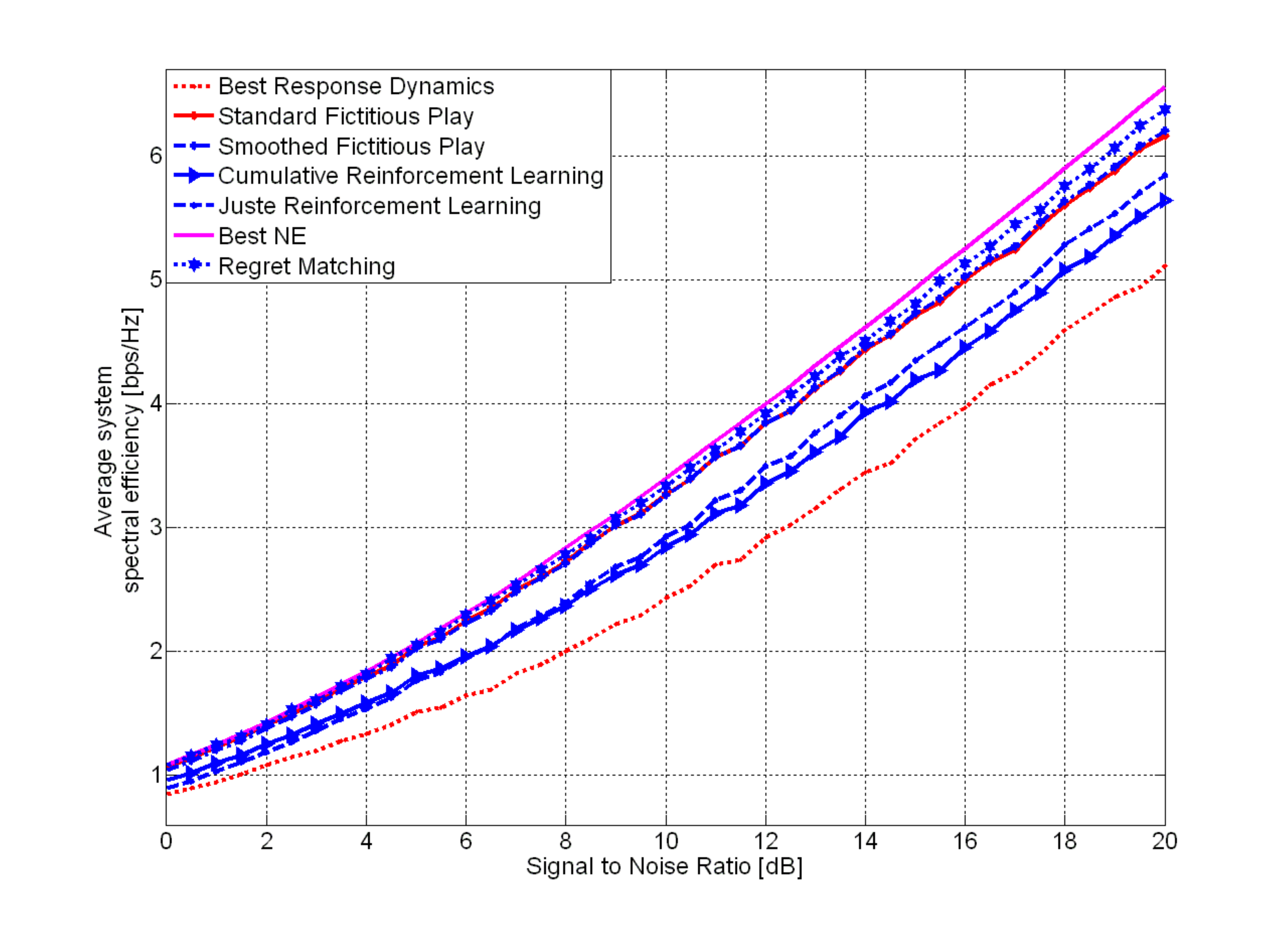}
\caption{Average system spectral efficiency [bps/Hz] as a function of signal to noise ratio (SNR) with $40$ iterations for the $2$ players and $2$ channel case.}
\label{fig:2x2perf_vs_iter}
\end{figure}

\begin{figure}[t]
\centering
\includegraphics[width=1\textwidth]{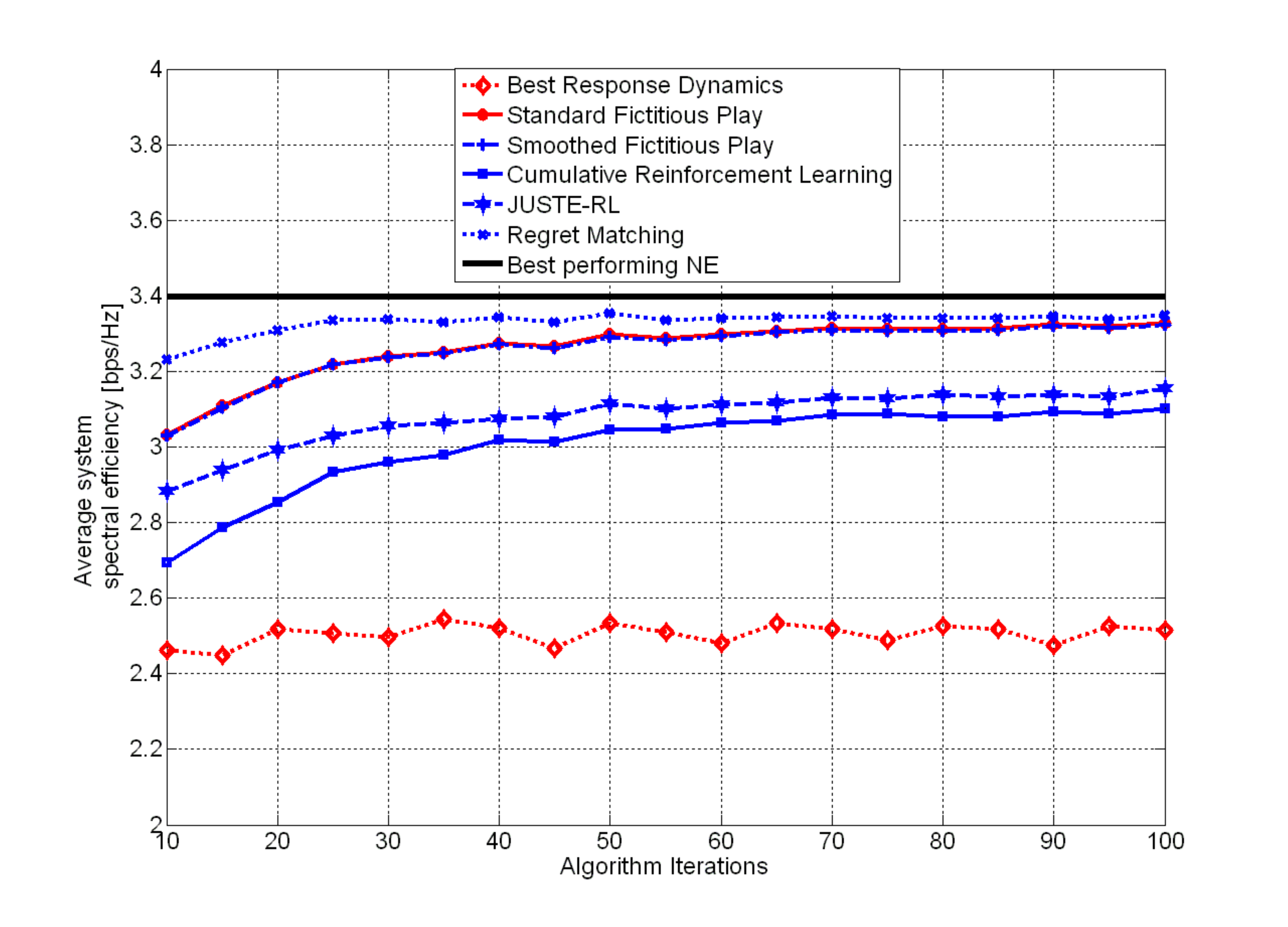}
\caption{Average system spectral efficiency [bps/Hz] as a function of the number of iterations at a fixed SNR of $10$ dB for the $2$ players and $2$ channel case.}
\label{fig:2x2perf_vs_SNR}
\end{figure}

\begin{figure}[t]
\centering
\includegraphics[width=1\textwidth]{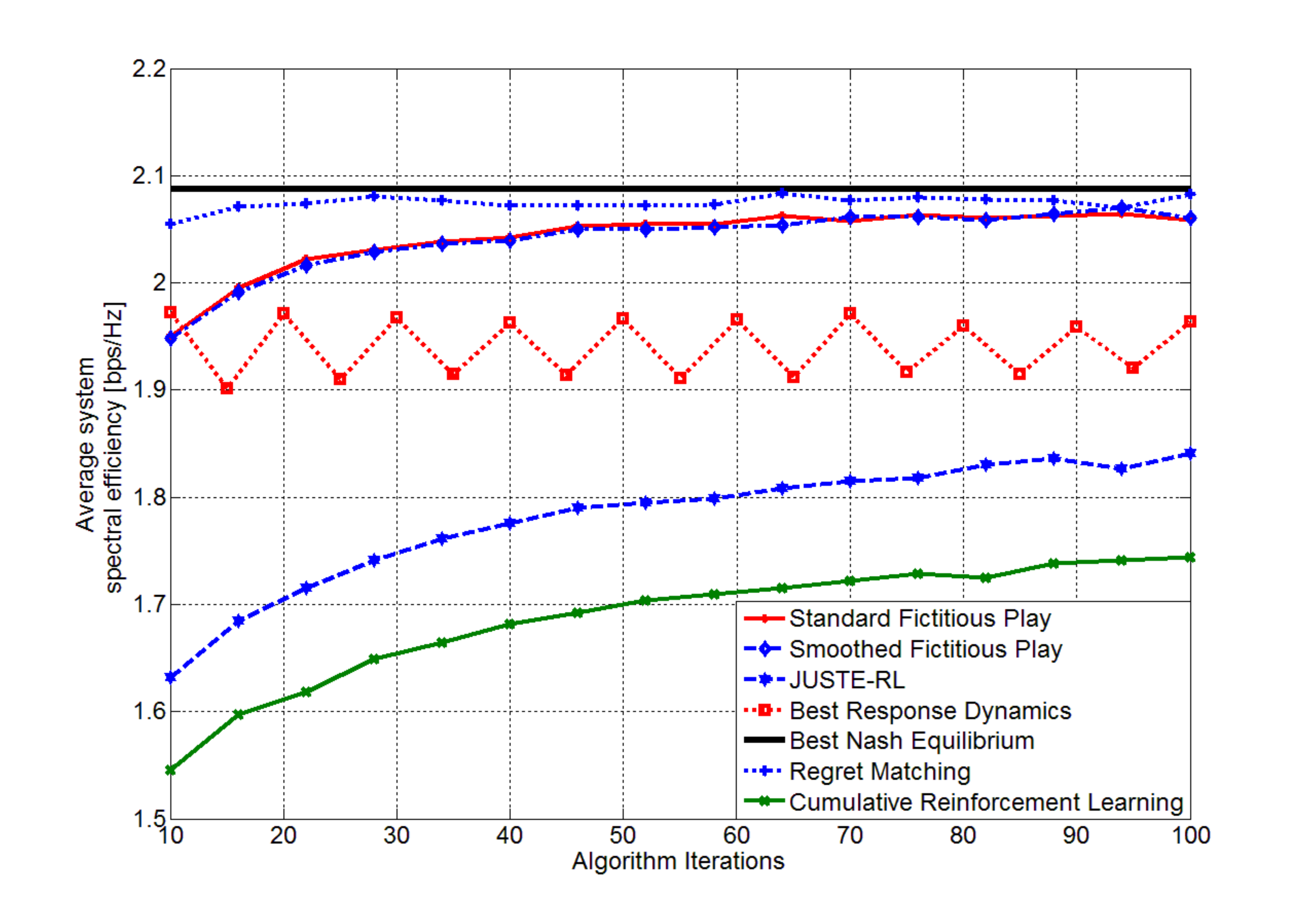}
\caption{Average system spectral efficiency [bps/Hz] as a function of the number of iterations at a fixed SNR of $10$ dB for the $2$ players and $4$ channel case.}
\label{fig:2x4_vs_iter}
\end{figure}

\noindent
In Figure \ref{fig:2x2perf_vs_iter}, we plot the average spectral efficiency of the system as a function of the SNR, in the case where only $2$ orthogonal channels are available. Here, all the algorithms iterate the same number of times ($40$ iterations).  In Figure \ref{fig:2x2perf_vs_iter}, it is interesting to note how algorithms such as FP, SFP and RM converge always very close to the best NE, i.e, the NE associated with the highest network spectral efficiency.  Nonetheless, this performance is achieved at the cost of a lot of information about the game. In particular, note that RL and JUSTE-RL are less performing, but at the same time, less demanding in terms of information. Interestingly, the BRD demands the same information assumptions than FP, SFP and RM. However, the performance is even worse that RL. This is due to the fact that BRD does not necessarily converge to a NE in this particular game.
In Figure \ref{fig:2x2perf_vs_SNR}, we plot the network spectral efficiency of the algorithms as a function of the number of iterations for the case of two channels. Here, 
RM and BRD appear to be the best performing and worst performing algorithms, respectively. With respect to the BRD, such a performance is due to a \emph{ping-pong} effect between two particular action profiles. In detail, since players are simultaneously selecting the channels with the highest gain, it may happen that the best channel is the same for both players. Thus, for instance, at odd iterations they both share the same channel and in the next one, they both select different channels. This effect will continue at the infinite preventing the algorithm to converge. 
In Figure \ref{fig:2x4_vs_iter}, we show how the algorithms perform when a higher number of channels is available, i.e, $4$ channels. BRD improves its performance, with respect to the other algorithms. This is mainly because the higher number of channel reduces the probability of the ping-pong effect described above. 

\begin{figure}[t]
\centering
\includegraphics[width=1\textwidth]{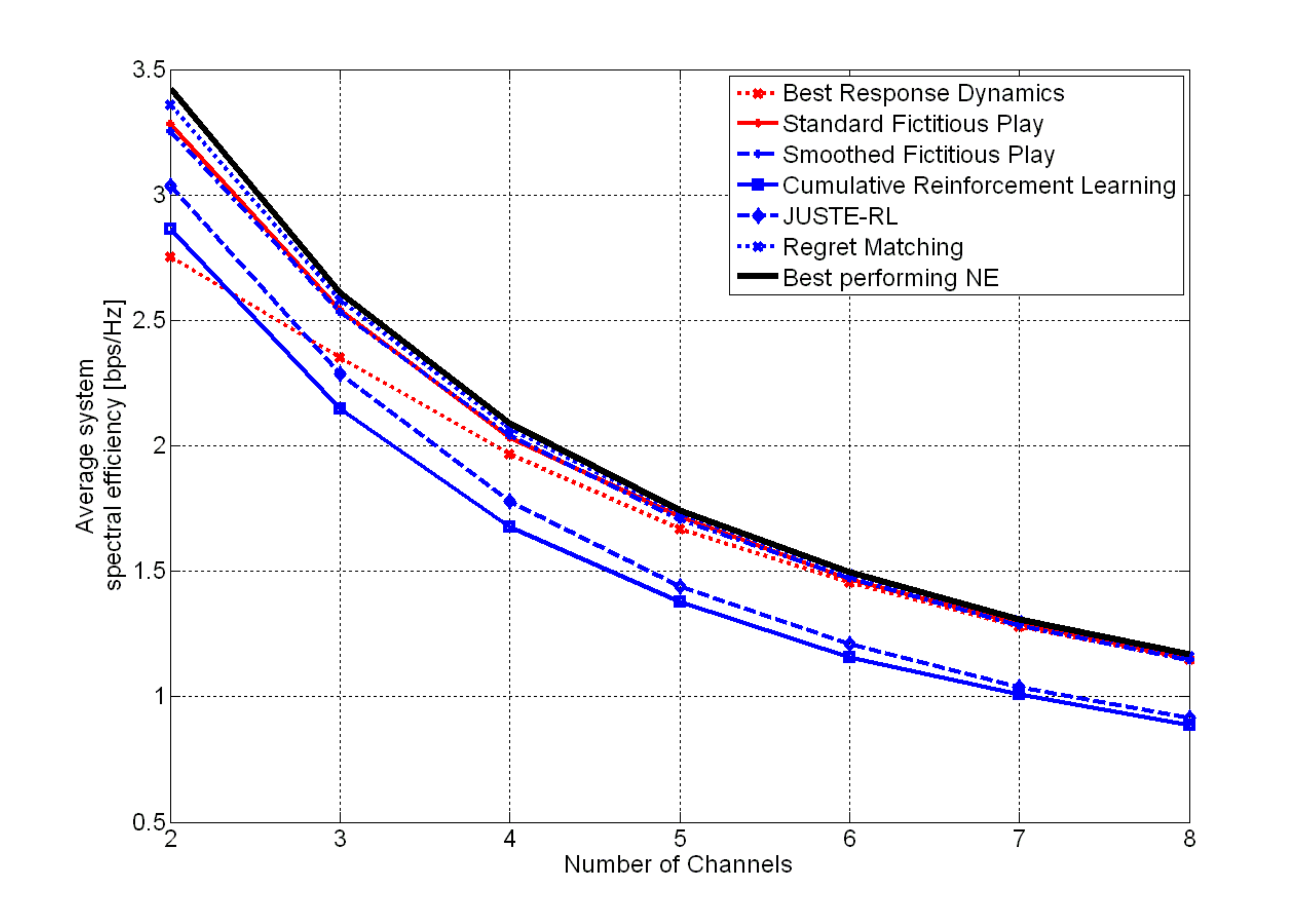}
\caption{Average system spectral efficiency as a function of the number of channels, with SNR=$10$dB and $40$ iterations.}
\label{fig:channels_varying}
\end{figure}

\noindent
In Figure \ref{fig:channels_varying}, we plot the network spectral efficiency as a function of the number of available channels. Here, the negative slot is due to the fact that we increase the number of available channels but transmitters remain subject to use a single channel. Thus, since $S > K$, there always exist a number of unused channels. The main observation in this figure is the following, the BRD becomes a very efficient solution when the number of channels is high enough to make the bouncing effect a very improbable event. Conversely, JUSTE-RL exhibits a lower performance when the number of possible actions increases. This is basically because, in JUSTE, all players play all their actions with non-zero probability in order to improve their utility estimation. Thus, this immediately implies that increasing the number of actions, increases the time that players are playing actions different from the optimal actions. 

\begin{figure}[t]
\centering
\includegraphics[width=1\textwidth]{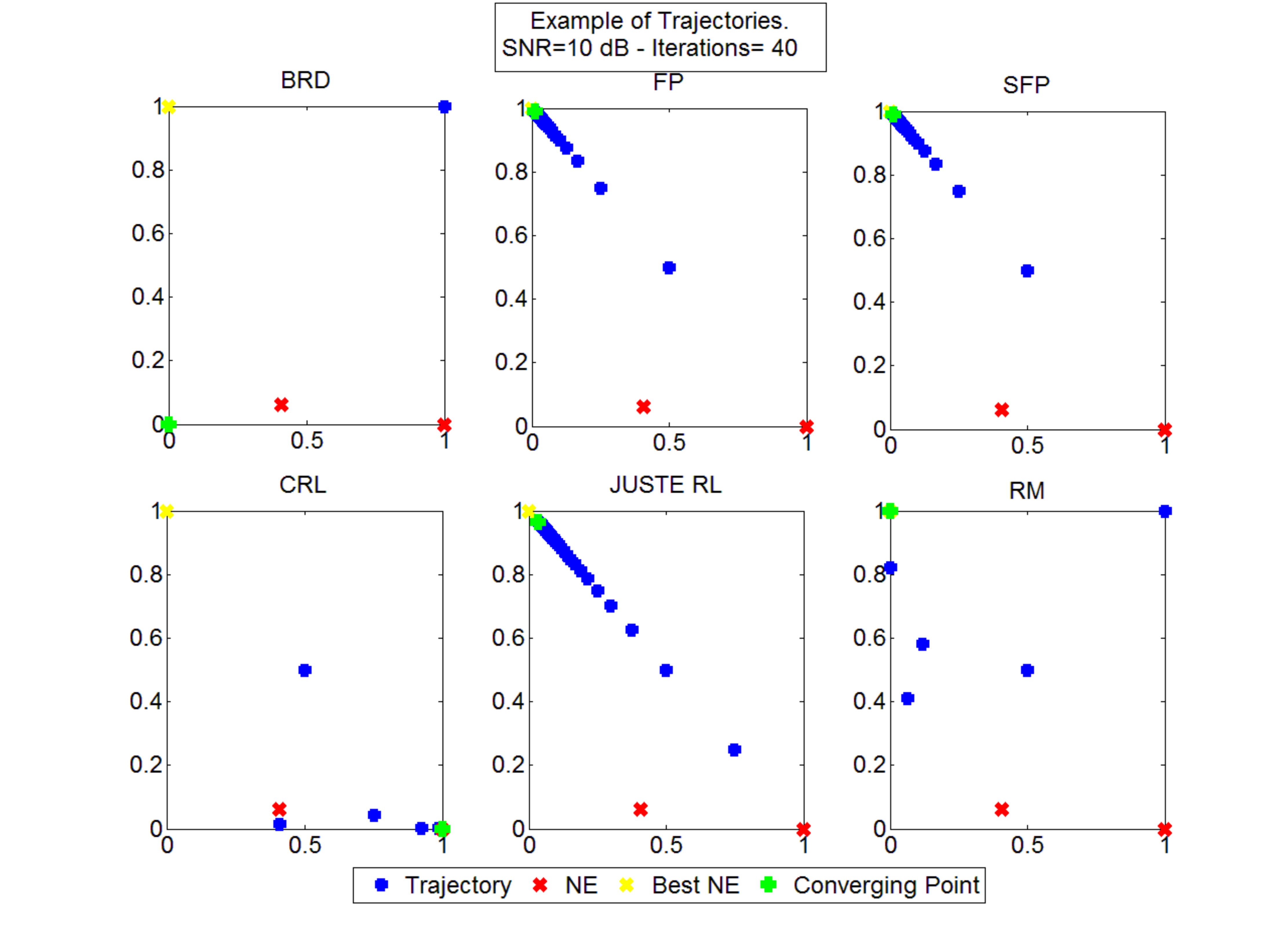}
\caption{Example of trajectories. BRD bounces between unstable solution; FP and SFP converge close to the best NE; CRL converges to a low performing NE, JUSTE-RL converges close to the best NE, RM converges close to the best NE}
\label{fig:Trajectories}
\end{figure}

\noindent
Finally, in Figure \ref{fig:Trajectories}, we show for the $2$-players $2$-channel case, the trajectories of the algorithm during the transient phase. In this realization, BRD it can be observed that BRD does not converge. The two transmitters repeatedly select synchronously the same channel. FP and SFP converge to the best performing NE while CRL converges fast to a steady point with no game theoretical meaning. In the trajectory of JUSTE, it is possible to see that it converges to the best performing NE, for that particular channel realization. Similarly, RM also converge very fast to the best NE.

\section{Conclusion}

In this paper, we have presented several notions of equilibrium and several learning dynamics that allow wireless networks to achieve such equilibria. In particular, we have described a general notion of equilibrium, namely, the coarse correlated equilibrium (CCE). Then, we introduced some particular cases of CCE, such as correlated equilibrium (CE) and Nash equilibrium (NE), are also analysed. Regarding the learning dynamics, we have presented the best response dynamics (BRD), fictitious play (FP), smooth fictitious play (SFP), regret matching (RM), reinforcement learning (RL) and joint utility and strategy estimation based reinforcement learning (JUSTE-RL). We have identified the pertinence of these algorithms for wireless communications in terms of system constraints (continuous/discrete actions, required information, synchronization, signalling, etc.) and the performance criteria (utility achieved at the steady state, convergence speed, etc.). As further work in this direction, we point out that existing results regarding the analysis of equilibrium in wireless networks strongly depend on the topology of the network. Indeed, a general framework for the analysis of equilibria and learning dynamics adapted to time-varying topology networks is still an open problem. Moreover, we must consider that some equilibrium notions, e.g., NE and $\epsilon$-NE, might be inefficient from a global point of view. Thus, learning algorithms to achieve Pareto optimal solutions with partial information is a further direction of research.

\bibliographystyle{IEEEtran}
\bibliography{GT}
\newpage

\begin{table}
\tiny
\centering
\begin{tabular}{|c|c|c|c|c|c|c|c|}
\hline
& BRD & FP & SFP & RM & RL & JUSTE-RL\\
\hline
Observations   & $\bs{a}_{-k}(t)$ & $\bs{a}_{-k}(t)$ & $\bs{a}_{-k}(t)$ & $\bs{a}_{-k}(t)$ & $\tilde{u}_k(t)$ & $\tilde{u}_k(t)$\\
\hline
Closed Expression for $u_k$ & Yes & Yes & Yes & Yes & No & No\\
\hline
Computation complexity & Optimization & Optimization & Optimization & Optimization & Algebraic Operation & Algebraic Operation\\
\hline
Steady State & NE & NE & $\epsilon$-NE & CCE & $--$ & $\epsilon$-NE\\
\hline
Condition for Convergence & DSG,PG, SMG & DSG, PG, ZSG,  $2\times N$-NDG & DSG,PG,ZSG & NDG & $--$ & DSG, $2-$player ZSG, PG\\
\hline
Synchronization to Play & Yes & Yes & Yes & Yes & No & No  \\
\hline
Environment & Static & Stationary & Stationary & Stationary & Stationary & Stationary\\
\hline
\end{tabular}
\caption{Benchmark of Learning Algorithms.}

\label{FigTable}
\end{table}

\end{document}